\documentclass[10pt]{article} 
\usepackage[preprint]{tmlr}

\usepackage{amsmath,amsfonts,bm}

\def\eqref#1{equation~\ref{#1}}

\def\1{\bm{1}}

\DeclareMathAlphabet{\mathsfit}{\encodingdefault}{\sfdefault}{m}{sl}
\SetMathAlphabet{\mathsfit}{bold}{\encodingdefault}{\sfdefault}{bx}{n}

\usepackage{hyperref}
\usepackage{url}
\usepackage{pstricks}
\usepackage{tikz}
\usepackage{graphicx}
\usepackage{wrapfig}
\usepackage{lipsum}
\usepackage{booktabs}
\usepackage[linewidth=0.6pt,linecolor=black]{mdframed}

\newcommand{\TK}{\textit{TopK}~}
\newcommand{\OT}{\textit{Odd-One-Out}~}

\title{Beyond Subtokens: A Rich Character Embedding for\\Low-resource and Morphologically Complex Languages}

\author{\name Felix Schneider \email felix.schneider@uni-jena.de \\
      \addr Computer Vision Group \\
      Friedrich Schiller University Jena
      \AND
      \name Maria Gogolev \email maria.gogolev@uni-jena.de \\
      \addr Computer Vision Group \\
      Friedrich Schiller University Jena
      \AND
      \name Sven Sickert \email sven.sickert@uni-jena.de \\
      \addr Computer Vision Group \\
      Friedrich Schiller University Jena
      \AND
      \name Joachim Denzler \email joachim.denzler@uni-jena.de \\
      \addr Computer Vision Group \\
      Friedrich Schiller University Jena}

\def\month{MM}  
\def\year{YYYY} 
\def\openreview{\url{https://openreview.net/forum?id=XXXX}} 

\begin{document}

\maketitle

\begin{abstract}
Tokenization and sub-tokenization based models like word2vec, BERT and the GPTs are the state-of-the-art in natural language processing.
Typically, these approaches have limitations with respect to their input representation.
They fail to fully capture orthographic similarities and morphological variations, especially in highly inflected and under-resource languages.
To mitigate this problem, we propose to computes word vectors directly from character strings, integrating both semantic and syntactic information.
We denote this transformer-based approach Rich Character Embeddings (RCE).
Furthermore, we propose a hybrid model that combines transformer and convolutional mechanisms.
Both vector representations can be used as a drop-in replacement for dictionary- and subtoken-based word embeddings in existing model architectures.
It has the potential to improve performance for both large context-based language models like BERT and small models like word2vec for under-resourced and morphologically rich languages.
We evaluate our approach on various tasks like the SWAG, declension prediction for inflected languages, metaphor and chiasmus detection for various languages.
Our experiments show that it outperforms traditional token-based approaches on limited data using \OT and \TK metrics.
\end{abstract}

\section{Introduction}

\begin{wrapfigure}{r}{6.7cm}
    \centering
    \vspace{-.47cm}
    \includegraphics[width=4.6cm]{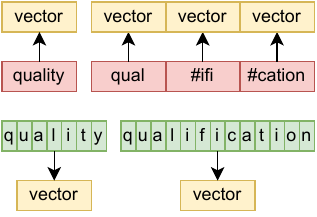}
    \vspace{-.27cm}
    \caption{An example for potential of subtokenization based embedding and our solution. We see the words first in a subtokenized manner and then with in our novel approach. The tokens that \textit{qualification} is split into are distinct to the token for \textit{quality}. Also, one word is split into three vectors. In contrast, our novel approach computes a single vector representation for each word and takes the spelling similarity into account.}
    \label{fig:teaser}
\end{wrapfigure}

As humans we have an intuitive understanding of words and their meaning.
For instance, we see the words \textit{quality} and \textit{qualification} as connected.
Even if we would not know the meaning of \textit{qualification}, we can infer its meaning by its similarity to the word \textit{quality} and the context it appears in.
However, in a pure dictionary-based input such a conclusion cannot be drawn.
Here, only context is available, while spelling similarities are not included.
Methods like WordPiece~\citep{devlin-etal-2019-bert} tokenization can mitigate this to some extent.
However, a standard English WordPiece model for BERT~\citep{devlin-etal-2019-bert} tokenizes \textit{quality} just in a single word, [quality].
At the same time, \textit{qualification} is tokenized into [qual], [ifi], and [cation].
It leads to the same problem of not having a concept of the similarity between \textit{qual} and \textit{quality} in the input.
A visualization of this example can be found in Figure~\ref{fig:teaser}.

Another challenge are different spellings.
The following deliberately mistyped sentence part highlights that \textit{tihs setnence can eeasily be raed by a hunam.}
A dictionary-based machine learning model on the other hand needs all writing variations - or at least enough of them - in the training corpus.
Otherwise a sensible vector representation for those words cannot be learned.

Finally, probably due to the nature of English as a default language used for science communication, dictionary-base approaches work well for this languages and related ones.
However, English is neither agglutinative nor highly inflected.
While there are also character- or byte-based approaches for context-based language models~\citep{xue2021byt5,clark2021canine,tay2021charformer}, this massively increases the length of the input sequence for the same text, since now istead of tokens or subtokens now every letter is input.
As a consequence, there is less context space available for the model.

Earlier LSTM-based approaches have a smaller context window and, thus, more quickly \textit{forget} previously provided information~\citep{khandelwal-etal-2018-sharp}.
Moreover, modern transformer-based approaches increase quadratically in memory requirement with their input sequence length~\citep{pmlr-v202-zhang23r}.
Naturally, the potential input sequence length is very limited.
For dictionary-based models a common mitigation strategy for spelling variations and uncommon words is to simply train on extremely large amounts of data~\citep{mikolov2013w2v}.
The hope is that the training corpus includes many spelling variations and contextual information for every variation of a word to gain a meaningful representation.

\begin{wrapfigure}{r}{6.7cm}
        \begin{mdframed}
        \textit{Norwegian:} \\
        Dette er en enkel eksempelssetning. \\

        \textit{Icelandic:} \\
        Þetta er einföld dæmisetning. \\

        \textit{Faroese:} \\
        Hetta er ein einfaldur dømisetningur. \\

        \textit{English:} \\
        This is a simple example sentence. \\

        \textit{German:} \\
        Dies ist ein einfacher Beispielsatz. \\
        \end{mdframed}

    \caption{A simple example sentence in various Germanic languages. The similarities between the languages are obvious. However, even the similar words would be treated as completely distinct tokens in a dictionary-based approach like WordPiece, without any information about their similarity in the input.}
    \label{fig:example}
\end{wrapfigure}

All these issues are amplified when dealing with low-resource and under-ressourced languages.
On the one hand, there are huge corpora for widely used languages like Mandarin, English, Spanish, French or German.
On the other hand, languages with a low amount of speakers like Faroese or dead historical languages, there exist only very small corpora of text~\citep{blaschke-etal-2023-survey,ross2018small}.
Furthermore, many of those languages for example in the Indo-European branch are highly inflected.
As a result, learning a representation for every inflected variation of a word is not feasible~cite{cotterell-etal-2016-morphological}.

Notably, languages that are related often also share vocabulary and spelling similarities, even if the exact spelling differs.
An example in Figure~\ref{fig:example} shows an excerpt of a text in various Germanic languages.
Of course, the similarities between the languages are striking.
Even for English with its strong influence from French and Latin, some words are still recognizable.
In a typical token- or subtoken-based approach all these words would still be encoded as completely distinct tokens without any information about their similarity in the input.

Given these shortcomings, we propose a different approach called \textit{Rich Character Embedding}.
Instead of a dictionary of tokens or subtokens, it uses a transformer-based neural network to compute a vector representation for a word based on its character string.
This vector can be directly used as a drop-in replacement for dictionary- and subtoken-based word embeddings in existing model architectures.

Our proposed representation fulfills multiple criteria.
First, words that frequently appear in similar context have similar vector representations, comparable to traditional dictionary-based approaches.
Additionally, words with similar spellings also yield similar vectors.
This is especially relevant for inflected variants or alternative spellings of a word.
Third, the representation also contains information about the spelling of the original character sequence to facilitate inference of grammatical features like case and number in inflected languages.
Moreover, even words that are rare or entirely absent from the training corpus can be encoded without resorting to subtokenization.
It leverages the learned representations of similarly spelled words.
Finally, the model is designed to be easily integrated as a plug-in input replacement for larger context-based language models such as BERT without significantly impacting their memory requirements.

\section{Related Work}
\label{sec:related}

For languages with large training corpora, vector-based text representations today are usually creted using large transformer models such as BERT~\citep{devlin-etal-2019-bert}, while for other tasks such as text generation GPT-like models are used~\citep{radford2018improving,radford2019language,brown2020language}. However, training such large models is challenging for under-resourced language due to limited data~\citep{wu-dredze-2020-languages}.
While multilingaugal variants like mBERT aim to address this, they tend to favour high-resource languages, often producing suboptimal tokenization and representations for low-resource languages~\citep{wu-dredze-2020-languages}.

Traditional word embeddings surch as word2vec~\citep{mikolov2013w2v} and FastText~\citep{bojanowski2017} provide an alternative that remains relevant for under-resourced languages due to their simplicity and lower data requirements~\citep{grave-etal-2018-learning}.
FastText in particular impoves over word2vec by incorporating subword information, which helps to mitigate issues with out-of-vocabulary words and morphologically rich languages.
However, both methods still require substantial corpora to learn meaningful representations, and their lack of contextual understanding limits their effectiveness~\citep{coto-solano-2022-evaluating}.

For static single-word embeddings, the field has explored convolution-based enhancements to word embeddings, incorporating subword structure and convolutional layers to improve representations~\citep{cao-lu-2017-conv}.
These methods aim to capture syntactic and morphological information more effectively while maintaining efficiency.
Despite these improvements, the evaluation of word embeddings in low-resource settings remains a challenge, as many standard benchmarks are designed for high-resource languages and often fail to reflect linguistic characteristics of under-resourced languages~\citep{coto-solano-2022-evaluating}.
The \OT and \TK metrics are designed to address this issue, providing a more nuanced evaluation of word embeddings that is better suited to low-resource languages~\citep{stringham-izbicki-2020-evaluating}.

The work of~\cite{neitmeier_2025_autoregressive} adresses the problem of tokenization.
They replace fixed subword tokenization with a byte encoder to create word embeddings and also a decoder to generate the next word byte by byte.
They match subword-tokenized baselines on downstream tasks but are more robust to spelling variations.

MrT5~\citep{kallini2025mrt5dynamictokenmerging} improves the byte-level sequence-to-sequence modeling of ByT5 by adding a learned token deletion gate to delete subsets of byte tokens.
By this the later layers use less tokens, which reduces the memore requirement and increases the speed. By this, the sequence length can be reduced by up to 75\%, while keeping similar performance.

One work that analyzes the problem of classic subtoken-based encoders for character awareness is the work by~\cite{cosma-etal-2025-strawberry}.
They analyzed why LLMs fail at simple tasks such as asking for the numbers of \emph{r}s in the word \emph{strawberry}.
They show that character-level understanding emerges very late and apruptly in training, a larger vocabulary size and larger token length resulting in later emergence.
They propose a character-aware module that additionally feeds character composition information into the following model.
Their experiments show earlier and more reliant learning of character reasoning with minimal cost.

There exist also various approaches to make the the aforementioned architectures more feasible for low-resource languages. These comprise methods like finetunening BERT for low-resource language understanding using active learning~\citep{griesshaber-etal-2020-fine}.
In contrast to those methods, our approach directly targets the input representation of the words for both word2vek-like and BERT-like models, making it a drop-in replacement for the traditional dictionary-based word embeddings. However, since the rest of the model architectures can remain the same, our approach can be combined with those methods to further improve the performance of the models on low-resource languages.

\section{Method Overview}
\label{sec:rce}

Our novel approach \textit{Rich Character Embedding} is based on computing a vector representation for words based on their character strings.
To achieve this, we represent every character as a one-hot-vector.
We use modifiers to represent special characters like capital letters or umlauts.

\subsection{Basic Definitions}
\label{sec:definitions}
For our approach, we define a training corpus $T$ as a collection of tokens $t_1, t_2, ..., t_k$ with $k$ tokens in the corpus.
Since we tokenize the text by whitespace separation, tokens represent the different words that are present in a corpus. We do not work with subtokens.
Every token consists of a character string $C(t_n) = (c_{n1}, c_{n2}, ..., c_{nl})$ with a length of $l$.
All characters $c$ belong to an alphabet dictionary $A$.
This dictionary also contains additional entries, as explained in section~\ref{sec:input}.
The embedding vector of the character string $C(t_n)$ is called $e(t_n)$

If we train with 1-hot-token prediction as explained in Section~\ref{sec:training}, we build a dictionary $D_T$ for the most common tokens in the corpus $T$.
This means that a subset of the tokens $t_n$ in the corpus have a dictionary entry $D_T(t_n)$.
Furthermore, $D_T$ is a subset of the set of all unique tokens in the training corpus.

\begin{figure}[t]
    \centering
    \includegraphics[width=.8\textwidth]{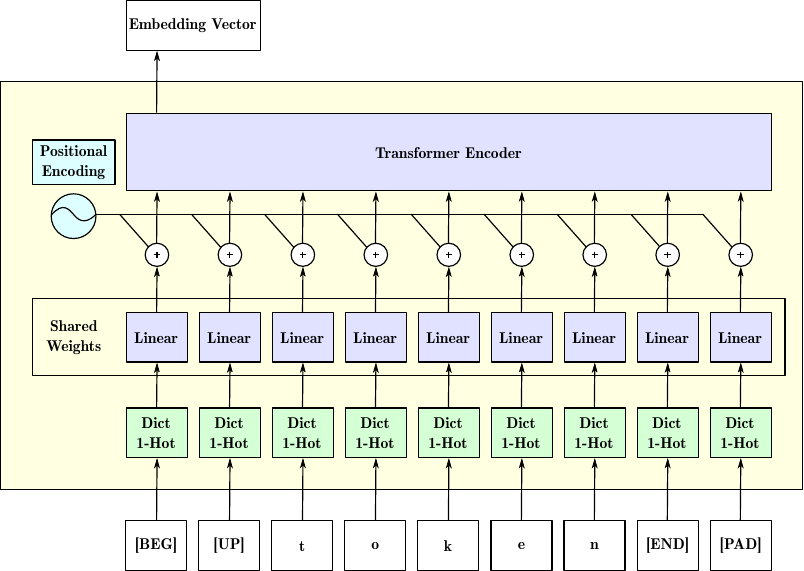}
    \caption{This figure shows the model architecture for the Rich Character Embedding. The input token, in this case the word \textit{Token} gets represented in the input as its character string, with the capital \textit{T} split up into an [UP] modifier and the base character \textit{t}. The character string is then transformed into a vector representation by the transformer encoder. The output of the encoder is then used as the word embedding.}
    \label{fig:modarch}
\end{figure}

We denote $E_T$ as the embedding model trained on the corpus $T$. The input of the model $E_T$ is the character string $C(t_n)$ instead of the dictionary entry of $D_T(t_n)$. That is, even for variations of words, such as different spellings, different inflections, suffixes and prefixes, etc. in the input, we are able to train the model.
If we train with 1-hot-token prediction instead of or additional to character string prediction, we need at least one word with a represenentation in a 1-hot-vector dictionary in its direct neighbourhood.
Figure~\ref{fig:modarch} shows a representation of the model architecture.

\subsection{Input Representation}
\label{sec:input}

In our character-level encoding text data is represented and processed at the granularity of individual characters. Traditional word-level or token-level encoding breaks text into words or subwords and produces a one dimensional vector for each. For character-level encoding the output is two dimensional for each word, as we create a one-hot vector for each character in the sequence.
The sequences begin with a [BEG] token and end with an [END] token.
The tokens are divided into the subgroups presented in Table~\ref{tab:tokens}.

\begin{table}[h]
    \centering
    \begin{tabular}{l|p{8cm}}
        \toprule
        \textbf{Category} & \textbf{Tokens} \\
        \midrule
        Base characters & a–z \\
        Digits & 0–9 \\
        Special symbols & \verb|ø|, \verb|þ|, \verb|ð|, \verb|ł|, \verb|ŋ|, \verb|°|, \verb|^|, \verb|!|, \verb|"|, \verb|§|, \verb|$|, \verb|%|, \verb|&|, \verb|/|, \verb|(|, \verb|)|, \verb|=|, \verb|?|, \verb|`|, \verb|´|, \verb|+|, \verb|*|, \verb|~|, \verb|#|, \verb|'|, \verb|_|, \verb|.|, \verb|:|, \verb|,|, \verb|;|, \verb|<|, \verb|>|, \verb|@|, \verb|€|, \verb|{|, \verb|[|, \verb|]|, \verb|}|, \verb|…|, \verb|„|, \verb|“|, \verb|”|, \verb|‚|, \verb|’|, \verb|—|, \verb|»|, \verb|«| \\
        Special word-level tokens & [BEG] (beginning of a word), [END] (end of a word), [UNK] (unknown character), [PAD] (padding token at the end of a word) \\
        Special character-level tokens & Padding and unknown characters \\
        Modifiers for diacritics & Acute Accent (\verb|´|), Grave Accent (\verb|`|), Macron (\verb|¯|), Tilde (\verb|~|), Horn (\verb| ˛|), Diaeresis (\verb|¨|), Ligature modifier, Uppercase modifier, Sharp S modifier (\verb|ß|) with base character 's' \\
        \bottomrule
    \end{tabular}
    \caption{Token categories and their corresponding characters.}
    \label{tab:tokens}
\end{table}

\subsubsection{Method}
Starting with a raw text sequence or sentence, we split the text into words, treating each word and special symbol as separate tokens.
Each word is converted character-by-character into its 1-hot-vectors representation.
For this we use a predefined vocabulary or character-token-to-index mapping.
Finally we pad each word to the length of the longest allowed word with a special character-level padding token.
Special symbols and special tokens such as [CLS], [SEP] and [PAD] are treated as one-character words.
For characters with diacritics, uppercase variations, or ligatures modifiers are used to encode them.
The word representation always starts with an additional [BEG] and ends with an [END] token, potentially followed by padding [PAD] tokens for batch processing.
When an uppercase character with a diacritic is encountered, it is encoded as follows: First, the appropriate modifier is added to indicate the presence of a diacritic. Then, if the character is in uppercase, the "up" modifier is applied to indicate that it's an uppercase character. Finally, the base character (the character without the diacritic) is encoded as if it were in lowercase. As an example: the word \textit{Liberté} would be encoded as [BEG] [UP] [l] [i] [b] [e] [r] [t] [´] [e] [END] [PAD], with the number of padding tokens being dependant on the maximum potential word length in a batch.
The principle of splitting up characters into multiple tokens is also applied to ligatures and other special characters.
The [END] token is added to the end of the word to indicate the end of the word for word reconstruction tasks for the training process.

In the following we give an example of the input representation for the word \textit{Token}.
\begin{enumerate}
  \item We split the word into it's characters: T, o, k, e, n
  \item We encode the capital T with an [UP] modifier and the base character t, while the other characters are encoded as they are: [UP] [t] [o] [k] [e] [n]
  \item We add the [BEG] and [END] tokens to the beginning and end of the word: [BEG] [UP] [t] [o] [k] [e] [n] [END]
  \item We pad the word to the maximum word length in the batch with [PAD] tokens: [BEG] [UP] [t] [o] [k] [e] [n] [END] [PAD] ... [PAD]
  \item We transform the character tokens into their one-hot-vector representation, resulting in a tensor of size $|A| \times |C|$ with $|A|$ being the size of the alphabet+modifier dictionary and $|C|$ being the length of the character string including the special tokens and padding.
  \item This tensor is then transformed via a linear layer to the embedding size $len(e) \times |C|$ and fed into the transformer encoder.
  \item The first output of the transformer encoder, which corresponds to the [BEG] token, is then used as the word embedding for the word \textit{Token} for downstream usage.
  \item During training, the output vector corresponding to the [BEG] token is used as an input for the word prediction heads expained below.
\end{enumerate}

\subsection{Model Architecture}
\label{sec:model}

Our Rich Character Embedding uses a classical transformer encoder model~\citep{vaswani2017attention} to generate the word embeddings.
The input is generated using the method described above and is a vector of the size $|A| \times |C|$ with $|A|$ being the size of the alphabet dictionary and $|C|$ being the length of the character string.
This is then transformed via a linear layer to the embedding size $len(e) \times |C|$.
In typical transformer model fashion, we then add a positional encoding vector to each of the $|C|$ input vectors.
In contrast to other transformer encoder applications, we only use the first output of the transformer encoder as the word embedding, the one that corresponds to the [BEG] token.
This corresponds to the common practice of using the [CLS] token output as the sentence embedding in BERT-like models.
Following this preparation, we feed the positionally encoded data to a transformer encoder.
The first output of the encoder is then used as word embedding.

\subsection{Model Training}
\label{sec:training}

Similar to traditional methods like word2vec and FastText, for every word in the corpus, we train the model by predicting the surrounding words.
Due to this training method the resulting word embedding vectors that are generated by the RCE model contain semantic information.
Additionally, we train the model to predict the character string of the input word itself.
Thus we ensure that also the syntactic information is preserved in the resulting word vector.

\subsubsection{Context Prediction}
\label{sec:context}

Similar to models like word2vec, we learn a vector represention model by predicting the surrounding words.
However, we do not predict dictionary entries of the tokens in this configurations, but their character strings.

We take the word vector $e$ from our embedding model and replicate it $l$ times, with $l$ being the maximum word length.
We then add positional embedding to those word vectors, in similar fashion to other transformer-based models.
The resulting tensor is then fed to another transformer encoder model. The output is then fed into a linear layer to predict the one-hot encoded characters of the surrounding tokens, similar to the input.
We decide for another encoder instead of a decoder model to be able to train the model end-to-end without the need to recursively predict character by character.
Fully achieving the training target of perfectly predicting the surrounding tokens is not necessary, similar to how classic word2vec models do not need to predict the surrounding words perfectly.
However, since we can use this method to train the model also on rare, unusually spelled, or unusually inflected surrounding words, the character-based context prediction enables our model to also learn from these words.
To summarize, the actual predicting performance is not important, but the training signal is.

\subsubsection{Identity Prediction}
\label{sec:identity}

Another way to train the model is to use the resulting word embedding vector $e$ to predict the encoded input word itself.
With this training goal, the model does not learn the semantic meaning of the word, but encodes the character string similar to an autoencoder.
This results in a word vector that also contains the syntactic information of the word, as opposed to just the semantic information.
The general training procedure is similar to the context prediction in Section~\ref{sec:context}.
This type of training is to be used in addition to a semantic type of training, to ensure that also the syntactic information is still preserved in the resulting word vector.

\subsubsection{Context Dictionary Prediction}

Word2Vec, FastText and even models like BERT work by predicting the dictionary entries of surrounding words or their sub-tokens.
While our approach does not use a dictionary-based approach to encode the words and generate the word vector, the training signal from traditional dictionary entry prediction is a simpler task than the character-based context prediction explained in Section~\ref{sec:context}, resulting in an easier training task for the model.
Thus, predicting the surrounding tokens - if they appear in the dictionary - solves as our third training objective, additional to the methods presented in Section~\ref{sec:context} and~\ref{sec:identity}.

\subsubsection{Hyperparameter Search}
\label{sec:hyperparams}

For the training heads mentioned in Section~\ref{sec:training}, we need to first find a set of hyperparameters for the encoder.
We used a limited training regime with only 100,000 training steps, with a batch size of 512.
To achieve good general hyperparemeters, we used a mixed corpus containing Icelandic, Norwegian, German, Latin, Faroese and Uzbek.
The hyperparameters we searched for were the learning rate (0.1, 0.01, 0.001, 0.0001, 0.00001), the embedding size (64, 128, 256),
the size of the transformer feedforward layer (64, 128, 256, 512), the number of encoder layers (2, 4), and the number of attention heads in the encoder (2, 4, 8).
We trained all combinations of the context prediction, identity prediction and context dictionary predictions from Section~\ref{sec:training}.
The chosen hyperparameters were a learning rate of 0.001, an embedding size of 64, 3 transformer encoder layers and 2 encoder attention heads.

\section{Evaluation}
\label{sec:eval}

Evaluating a word embedding model is different from evaluating models that are directly trained for tasks such as classification.
Two distinct methodologies may be used:
1. analyze the properties of the embedding vectors to ascertain their alignment with specified criteria, and 2. measure the performance of the word embedding vectors on downstream tasks.
Since analyzing the properties is problematic in the context of low-resource languages, we analyze the properties using the \OT and the \TK methods~\citep{stringham-izbicki-2020-evaluating}.
As downsteam tasks we employ varous experiments like sentence similarity tasks and declension prediction for inflected languages.

\subsection{Experimental Setup}
\label{sec:desc}
We pretrain our models on the CC100 corpus~\citep{conneau-etal-2020-unsupervised}, besides Uzbek, which we train on an Uzbek news corpus~\citep{kuriyozov_elmurod_2023_7677431}.

\subsection{Baseline Methods}
For the baseline methods, we use FastText and char2vec.
As described above, FastText is an extension of word2vec that incorporates subword information.
Since many modern static word embedding approaches also use convolution-based character information, we use a simple implementation similar to char2vec~\citep{ho2025char2vec}, which we denote with c2v.
The character string representations we use for the char2vec model are the same as for the RCE model.

\subsubsection{Metrics}
We use \TK and \OT to analyze (i) whether our embedding vectors have the desired property that words that appear in similar contexts have similar embedding vectors and (ii) words that appear in different contexts have distinct embedding vectors.
These two metrics are designed to especially analyze low-resource languages~\citep{stringham-izbicki-2020-evaluating}.
Both require having a dataset with categories and members of these categories.
For example, the category \textit{transportation} could contain elements like \textit{bus}, \textit{car}, and \textit{plane}, while \textit{politics} could contain \textit{chancellor}, \textit{president}, and \textit{parliament}.

For the TopK metric, we compute the word vectors of all the words in our evaluation dataset.
For every words we search the top $k$ - with $k=3$ in our case - closest words, in terms of euclidean distance of their word vectors.
The percentage of closest words in the same category is then the score for the word.
The TopK score for the whole dataset is the average value over all words. A score of 1 implies that in the word vector space every word is surrounded just by words of the same category.
In contrast, a score of 0 indicates that every word is clustered with words of different categories.

For the OOO metric, we choose a random set of 10 words as indicated in~\citep{stringham-izbicki-2020-evaluating} from one category and one word from a random different category. We compute the mean word vector of those 11 words and then choose the one that is furthest away from this prototypical vector. If this word is the outlier from the different category, we assign the set a value of 1, otherwise 0.
Then, the OOO score is the average value over a number - in our case 1000 - different random sets.

\begin{table}
  \centering
\begin{tabular}{lrrr}
\toprule
language & categories & total words & average words per category \\
\midrule
Faroese & 8 & 161 & 20 \\
Latin & 9 & 158 & 18 \\
Uzbek & 9 & 141 & 16 \\
\bottomrule
\end{tabular}
  \caption{This table shows the statistics of the datasets we created for the TopK and OddOneOut evaluation methods.}
  \label{tab:dataset}
\end{table}

We have constructed datasets with words for different categories for Latin, Faroese, and the Uzbek language containing categories and category members. You can see the statistics for dataset in Table~\ref{tab:dataset}.

\subsubsection{Inflection}
In contrast to English and Mandarin Chinese, many languages such as Latin are highly inflected.
This means that e.g. the appearance of nouns change depending on the casus und numerus of the word.
Since this means that for every form of a word a separate dictionary entry is needed for traditional methods - at least if tokenization like WordPiece does not spearate the stem from the ending by chance - the representation of even the base form should be worse than with a method that considers also the similar word stems, like our novel approach.
Additionally, the information contained in the inflected word endings should carry over to other words that have not yet been seen beforehand.
If a human knows how a certain declination behaves in a language, they can infer differently declined forms from just the base form of the word or infer the base form from a declined one.
The same holds true for other forms of inflection, like conjugation.

To evaluate how our novel method fares with inflection, we have constructed two new tasks for the Latin language:
\textbf{Predicting the casus and numerus} of nouns and \textbf{predicting their declension} from the nominative and genitive singular of a noun.
We created a dataset consisting of 10,000 Latin nouns of various declensions.
The dataset was generated by crawling a Latin dictionary website\footnote{\url{https://www.latin-is-simple.com}} and choosing a random subset of words.
We incorporate in the dataset every variation of a word in terms of casus and numerus, resulting in 12 forms per word, 6 singular and 6 plural.
This results in a set of 12,000 samples, with 10,000 samples per class.

\subsubsection{Stylistic Device Detection}
In the field of stylometry, \textit{Stylistic Device Detection} is a field that has been studied for long in the case of some stylistic devices like e.g. \textit{metaphors} and only studied in a very limited fashion in the case of the \textit{chiasmus}.
Since both of the beforementioned stylistic devices have also been analyzed in low resource languages which posed additional problems, we think that a comparison of our word vectors with the word vectors used in those studies can highlight the performance of our novel approach.

\paragraph{Chiasmus Classification}
For the chiasmus classification we use the current state-of-the-art method presented in~\citep{schneider-etal-2021-data}.
To classify whether a phrase constitutes a chiasmus or not, the authors use various features, including pairwise distances of word embeddings the four main words constituting the chiasmus, resulting in 6 vector distances as features.
In our experiment we use both our novel word embedding approach and compare it to our baselines. Since this method has also been applied to Middle High German, which is a historical low-resource language, we use their dataset to evaluate the methods. For this we split the dataset into 5 parts and ran a 5-fold crossvalidation.

\paragraph{Metaphor Classification}
For the metaphor classification we use the method for low-resource metaphor prediction that was presented in~\citep{schneider-etal-2022-metaphor}.
Since this method is especially geared towards low-resource languages and has also already been evaluated on Middle High German similar to their chiasmus classification, we replace the word embeddings used in their work with our novel approach and the baseline methods.
We again split their dataset into 5 parts and ran a 5-fold crossvalidation.

\subsubsection{Transformer-Based Contextual Embeddings}

For comparison, we trained both a BERT-like model with traditional WordPiece tokenization and a BERT-like model with our novel approach.
The [CLS], [SEP], [PAD], and [MASK] tokens that are used in the BERT model are treated as normal words consisting of their characters in this approach.
We then evaluated the models on a downstream experiment similar to SWAG~\citep{zellers2018swagaf}, but with datasets we constructed out of corpora of the languages. We make the datasets available together with our code.

As hyperparameters for the model we chose feedfoward layer/general tensor size of 128, 4 attention heads, and 4 encoder layers for the language transformer. For the character-based version we used an embedding size of 128 and two layers both for the character token encoder and decoder.
The batch size is 12 sentence pairs per batch.
Every sentence pair consists of two sentences from the training corpus, with the second sentence being either random or the next sentence in the corpus, and a label indicating whether the second sentence is a continuation of the first sentence or not.
We train both model types using next sentence prediction and masked language modeling and we replace 15\% of the tokens in the input with a [MASK] token.
For optimization, we use the Adam optimizer with a maximum, learning rate of 0.001, mediated by a linear warmup over the first 5000 steps and a cosine annealing schedule afterwards, for a total of 300000 steps.
The dropout rate is set to 0.1.

\subsection{Results}
\label{sec:results}

\subsubsection{TopK and Odd-One-Out}

\begin{table}
  \centering
  \begin{tabular}{r r c c c c}
      \toprule
      target language & training language & RCE & c2v & RCE+c2v & FastText \\
      \midrule
      Faroese & Faroese   & 0.26 & 0.20 & 0.25 & 0.21 \\
      Faroese & Norwegian & 0.22 & 0.20 & 0.24 & 0.24 \\
      Faroese & Icelandic & 0.24 & 0.22 & 0.26 & 0.22 \\
      \midrule
      Latin & Latin       & 0.23 & 0.25 & 0.27 & 0.25 \\
      Latin & Italian     & 0.16 & 0.26 & 0.24 & 0.15 \\
      \midrule
      Uzbek & Uzbek       & 0.20 & 0.25 & 0.25 & 0.20 \\
      \bottomrule

  \end{tabular}
  \caption{This table shows the results for the TopK evaluation. The scores are bounded between 0 and 0, with 1 being a perfect score and 0 being the worst score. The experiments were conducted on a Faroese dataset, using models trained on either Faroese, Icelandic, or Norwegian. We compare the results of RCE, c2v, and FastText. Since the whole testing dataset was used without any training splits, we can not provide standard deviations.}
  \label{tab:topk}
\end{table}

\begin{table}
  \centering
  \begin{tabular}{r r c c c c}
      \toprule
      target language & training language & RCE & c2v & RCE+c2v & FastText \\
      \midrule
      Faroese & Faroese   & 0.13 & 0.12 & 0.29 & 0.08 \\
      Faroese & Norwegian & 0.22 & 0.29 & 0.24 & 0.09 \\
      Faroese & Icelandic & 0.11 & 0.14 & 0.16 & 0.09 \\
      \midrule
      Latin & Latin       & 0.25 & 0.60 & 0.30 & 0.10 \\
      Latin & Italian     & 0.18 & 0.35 & 0.31 & 0.08 \\
      \midrule
      Uzbek & Uzbek       & 0.17 & 0.36 & 0.39 & 0.10 \\

      \bottomrule

  \end{tabular}
  \caption{This table shows the results for the OddOneOut evaluation. The scores are bounded between 0 and 0, with 1 being a perfect score and 0 being the worst score.}
  \label{tab:ooo}
\end{table}

We manually created a dataset for the Faroese, Latin and Uzbek languages, containing multiple categories and on average 20 members per category for Faroese, 18 for Latin and 16 for Uzbek.
We then trained models on the Faroese, Icelandic, and Norwegian datasets for Faroese, on Latin and Italian for Latin and on Uzbek for Uzbek and evaluated them on the Faroese dataset.
The results of the TopK and OOO experiments are shown in Table~\ref{tab:topk} and Table~\ref{tab:ooo}.
Our training approach for RCE and c2v always outperforms FastText for both metrics.
The best results are achieved by training RCE and c2v together in a combined model.
For TopK the metrics show a clear tendency for our RCE approach, both alone and in combination with c2v.
In case of the models trained on the Icelandic dataset, the RCE based approaches even outperform FastText trained on the target language, the Faroese datset.
For the OOO metric, the results show an even better performance of the character-based models.
The combined approach containing RCE and c2v outperform the other models by a large margin.
For RCE and c2v on their own, c2v slightly outperforms RCE.
These results show that our novel approach is able to generate word embeddings that are better aligned with the desired properties than traditional methods like FastText and thus more suitable for the generation of word vectors, especially in low-resource languages.

Since RCE+c2v generally perform similar to or better than other models trained on the target language, even when not trained on the target language, we can infer that our approach is able to capture the semantic information of the words better than traditional methods like FastText.

\subsubsection{Declension Detection by Inflection}

\begin{table}
  \centering
  \begin{tabular}{r c c c c}
    \toprule
    training language & RCE & c2v & RCE+c2v & FastText \\
    \midrule
    Latin & $0.85\pm0.02$ & $0.90\pm0.01$ & $0.90\pm0.01$ & $0.85\pm0.02$ \\
    Italian & $0.85\pm0.01$ & $0.88\pm0.02$ & $0.84\pm0.01$ & $0.50\pm0.01$ \\
    \bottomrule
  \end{tabular}
  \caption{This table shows the results for the inflection experiments. The experiments were conducted on a Latin dataset. We compare the results of RCE, c2v and FastText. The cores are bound between 0 and 1, with 1 being a perfect score and 0 being the worst score.}
  \label{tab:latininf}
\end{table}

To evaluate the performance of our novel approach in inflected languages, we conducted two experiments on the Latin and Italian languages.
We created a dataset for the Latin language from a dictionary, categorizing 18497 words into their declensions.
Since the declension of a word can be inferred from the nominative and genitive singular of a noun, we used those two forms as input for a simple feedforward neural network model with one hidden layer that predicts the word declension from the two concatenated vectors of the input words.
We then compared the results of the RCE model with the c2v model and FastText.
The results are shown in Table~\ref{tab:latininf}.

The results show that the c2v model with our novel training method and the RCE model perform both well when pretrained on a Latin and an Italian corpus. Both models consistently outperform FastText in both languages. c2v slightly outperforms RCE on this task.
However, it is notable that FastText performs better when pretrained on a corpus in the target language Latin and strongly decreases in performance when pretrained on a different language like Italian.
This indicates that our novel approach is able to capture the inflectional information of the words better than traditional methods like FastText.

\subsubsection{Stylistic Device Detection}

For the stylistic device detection, we tried the detection of metaphors and the detection of chiasmi.

\textbf{The chiasmus} is a stylistic device which works by following a phrase with its inversed version.
Often chiasmi are used to emphasize the contrast between the contents of the first and the second part of the phrase.
An example would be the phrase \textit{\textbf{long} is the \textbf{research}, but the \textbf{paper} is \textbf{short}}, emphasizing the contrast betwine the huge amount of research and the short possible length of the resulting paper.
\citep{schneider-etal-2021-data} published a method to detect chiasmi. This method is based on various features, among them the pairwise cosine distances of the word embeddings of the four main words constituting the chiasmus, in our case of the vectors for \textit{long}, \textit{research}, \textit{paper}, and \textit{short}.

\begin{table}
  \centering
  \begin{tabular}{r c c c c}
    \toprule
                  & RCE & c2v & RCE+c2v & FastText \\
    \midrule
    all features   & $0.34\pm0.17$ & $0.32\pm0.13$ & $0.30\pm0.11$ & $0.31\pm0.13$ \\
    only embedding & $0.19\pm0.06$ & $0.02\pm0.00$ & $0.32\pm0.14$ & $0.02\pm0.00$ \\
    \bottomrule
  \end{tabular}
  \caption{Results for the chiasmus detection experiments. We compare the results of RCE, c2v, and FastText. The values are bounded between 0 and 1, with 1 being a perfect score and 0 being the worst score.}
  \label{tab:chiasmi}
\end{table}

We trained our RCE, c2v, and FastText models on the German corpus and evaluated them on the German dataset.
For the train/test split we used a 5-fold crossvalidation.
Table~\ref{tab:chiasmi} shows the results of the chiasmus detection experiments.
When we use the full feature set from \citep{schneider-etal-2021-data}, we can see that all models perform equally well on the task.
However, when we remove all other features besides the embedding features to directly measure their impact, the mixed RCE+c2v model performs best, followed by our RCE model.
Fasttext and c2v alone perform much worse than our approach.
However it is interesting, that the combination of the convolutional and transformer-based approaches yields such a high improvement in performance compared to the transformer-based approach alone, especially when we look at how unsuitable the convolutional approach is on its own for this task.
This indicates that our novel approach and the convolutional approach can encode different information in the word embeddings, which can be beneficial for downstream tasks.

\textbf{The metaphor} is a stylistic device wich consists of using a word or a phrase from one domain in another domain to create a new meaning. To imply good weather, one could write \textit{the sun is smiling} - it is clear for the reader, that the sun does not actually smile, but produce warmth in a pleasant way. However, there are many different forms of metaphors. \citep{schneider-etal-2022-metaphor} published a method to detect adjective-noun metaphors like \textit{moody weather} or \textit{thirsty car}.
This method works by transforming the vector representation of both the adjective and the noun into a new metaphoricity space.
In this space, the cosine distance between the adjective vector and the noun vector is a measure of the metaphoricity of the phrase.
Both vectors are transformed by the same model.

\begin{table}
  \centering
  \begin{tabular}{ c c c c }
    \toprule
     RCE & c2v & RCE+c2v & FastText \\
    \midrule
     $0.73\pm0.05$ & $0.71\pm0.05$ & $0.74\pm0.04$ & $0.67\pm0.04$ \\
    \bottomrule
  \end{tabular}
  \caption{This table shows the results for the metaphor detection experiments. We compare the results of RCE, c2v, RCE+c2v, and FastText. The results are for the German dataset. The values are bounded between 0 and 1, with 1 being a perfect score and 0 being the worst score.}
  \label{tab:metaphors}
\end{table}

Table~\ref{tab:metaphors} shows the results of the metaphor detection experiments.
Similar to the previous experiments we trained RCE, c2v, a combination of both methods and FastText on the German corpus.
We then used the German dataset to evaluate the models.
To do this, we did a 10-fold crossvalidation.
It can be seen that the character-based models always perform better than FastText.
Additionally, our RCE model always performs better than the c2v model.
The highest performance is trained by the combined RCE+c2v model.

\subsubsection{BERT-like Model}

\begin{table}
  \centering
  \begin{tabular}{r r r c c }
    \toprule
    pretraining & task training & testing & WordPiece & RCE \\
    \midrule
    en & en & en & 0.561 & \textbf{0.578} \\
    de & de & de & 0.542 & \textbf{0.628} \\
    \midrule
    fo & fo & fo & \textit{0.279} & 0.265 \\
    no & no & fo & 0.277 & \textbf{0.285} \\
    no & fo & fo & \textit{0.266} & 0.256 \\
    is & is & fo & 0.244 & \textit{0.280} \\
    is & fo & fo & \textit{0.275} & 0.241 \\
    \midrule
    no & no & no & 0.537 & \textbf{0.561} \\
    no & fo & no & \textit{0.282} & 0.253 \\
    \midrule
    is & is & is & 0.353 & \textbf{0.391} \\
    is & fo & is & 0.274 & \textit{0.333} \\
    \bottomrule
  \end{tabular}
  \caption{This table shows the results for the SWAG-like next sentence prediction task. The values are bounded between 0 and 1, with 1 being a perfect score and 0 being the worst score. In this case Word Piece and RCE were used as an embedding layer for a BERT-like model.}
  \label{tab:swag}
\end{table}

Table~\ref{tab:swag} shows the results of the BERT-like model and the  model on the SWAG-like task.
In the first part we compared English, German, Norwegian and Icelandic.
We first pretrained a model on the respective language.
In the then finetuned it for the task on the same language, and finally evaluated it on the same language.
In all of those four cases, the our novel  model outperformed the BERT-like model we compared our appraoch with.

In the second part, we test the models on the low-resource Faroese language. We pretrain the model on either Faroese, Norwegian, or Icelandic. We then finetune the model on the SWAG-like training set in either Faroese target language or the language of the pretrained model. Finally, we evaluate the model on the Faroese test set.
The best results on the Faroese test set were achieved with both pretraining and finetuning on Norwegian with our novel  approach, while the second best results were achieved with both pretraining and finetuning on Icelandic, also with our novel RCE BERT-like approach.

This experiment indicates that our novel approach is also useful in the context of larger models like BERT, and especially useful in the context of low-resource languages, when pre-training on a different corpus is needed.

\section{Limitations}
\label{sec:limit}

Our novel approach computes word embeddings directly from character strings.
It has shown promising results for languages with alphabet-based writing systems.
However, the current method is developed for linear, one-dimensional writing systems such as those based on the Latin alphabet.
This limits its applicability to languages with non-alphabetic or more complex writing systems - such as Mandarin Chinese or Japanese - where characters are constructed from multiple subcomponents arranged in two dimensions.
While the underlying principles of character-based encoding may be adaptable to these systems, additional work is required to appropriately capture their structure.

The present implementation relies exclusively on a predefined set of Latin characters and associated diacritics.
This, it does not directly support other alphabet-based systems such as Cyrillic, Devangari, or Hangul.
To extend the model to encompass these scripts necessitates a redesign of the character-token mapping and the encoding step.

Finally, while our experimental evaluations indicate that the approach is effective within the tested contexts, the scope of these experiments is limited by both the range of languages and datasets included and by the constraints of available training hardware.
Future research should aim to scale up evaluations across a broader array of languages and leverage more extensive computational resources to fully explore the ability of our method to generalize and the performance on more languages.

\section{Conclusion and Outlook}
\label{sec:conclusion}

In this work we proposed a new method for generating word embeddings for low-resource languages.
We name this method \textbf{Rich Character Embedding} (RCE).
The method is based on the idea of generating word embeddings from the character string of a word.
We have shown in various experiments that our approach outperforms traditional methods like FastText in various tasks.
Furhtermore, we also have shown that our novel approach can be used as a drop-in replacement for traditional word embeddings in larger models like BERT.
Especially for low-resource languages, our approach is able to generate word embeddings that are better aligned with the desired properties of capturing both semantic and syntactic information and also be trainable on small corpora than traditional methods like FastText or sub-tokenization methods like WordPiece.

\bibliography{bibliography}
\bibliographystyle{tmlr}

\end{document}